\documentclass[letterpaper, 10 pt, conference]{ieeeconf}  

\IEEEoverridecommandlockouts                              

\usepackage{amsmath}
\usepackage{amssymb}
\usepackage{graphicx}
\usepackage{booktabs}
\usepackage{multirow}
\usepackage{makecell}
\usepackage{diagbox}
\usepackage{url}

\usepackage{booktabs}
\usepackage{tabularx}

\usepackage[
  ruled,
  linesnumbered,
  boxed,
  noend
]{algorithm2e}

\graphicspath{{figures/}}

\title{\LARGE \bf
PerchRL: Vision-Based Agile Perching on Inclined Platforms under Rapid and Irregular Motion
}

\author{Zihong Lu$^{1,2,*}$, Zongzhuo Liu$^{1,*}$, Huaxu Li$^{1}$, Yitao Zeng$^{1}$, \\ Jinqiang Cui$^{3}$, Jie Mei$^{2}$, Youmin Gong$^{2}$, U Kei Cheang$^{1}$, Boyu Zhou$^{1,4,\dagger}$ 
\thanks{$^*$ indicates equal contribution. $^\dagger$ indicates Corresponding Author.}
\thanks{$^{1}$Southern University of Science and Technology, Shenzhen, China}%
\thanks{$^{2}$Harbin Institute of Technology, Shenzhen, China}%
\thanks{$^{3}$Peng Cheng Laboratory, Shenzhen, China}%
\thanks{$^{4}$Differential Robotics}%
\thanks{Email: luzong2001@gmail.com, zhouby@sustech.edu.cn}%
}

\begin{document}

\maketitle
\thispagestyle{empty}
\pagestyle{empty}

\begin{abstract}
    Autonomous vision-based perching of quadrotors on moving inclined platforms is critical for air-ground collaboration but remains challenging due to the limited field of view (FOV). In this paper, we propose PerchRL, a reinforcement learning (RL) framework for vision-based agile perching on inclined platforms under rapid and irregular motion. Specifically, we employ a two-stage learning strategy consisting of state-based pre-training followed by vision-based fine-tuning. To improve generalization across diverse platform motions, we employ randomized platform trajectories to prevent overfitting and temporal augmentation methods to capture latent motion patterns from historical observations. During vision-based fine-tuning, a hybrid learning framework consisting of visibility-aware state augmentation and active perception rewards is presented to improve robustness under intermittent visual loss. Extensive simulation and real-world experiments demonstrate the feasibility, stability, and real-time performance of PerchRL, while successful deployment across distinct quadrotor platforms further validates its adaptability. The source code will be released to benefit the community.
\end{abstract}

\section{Introduction}

Autonomous perching of quadrotors on moving inclined platforms is critical for air-ground collaboration applications \cite{das2020synchronized,wu2022collaborative}, such as mobile recharging and emergency response. Vision-based approaches \cite{krogius2019flexible,wen2020se} serve as a widely used perception modality for aerial perching, providing flexible and lightweight relative pose estimation between the drone and the target platform. However, the limited FOV, underactuated dynamics, and aggressive maneuvers inherent to agile perching make intermittent visual loss inevitable in practice, challenging the continuous-feedback assumption of most existing methods.

Traditional vision-based perching methods explicitly predict platform motion and perform trajectory planning via multi-constrained optimization, resulting in high computational overhead and limited replanning frequency. Furthermore, this tight perception–prediction–planning coupling strongly depends on accurate motion prediction, which becomes unreliable under complex platform motion, leading to accumulated closed-loop errors. Consequently, end-to-end RL-based approaches have recently been explored, enabling high-frequency control with reduced system coupling. However, existing RL methods still suffer from two critical limitations. First, most approaches rely on fixed or highly structured trajectories during both training and evaluation, leading to overfitting and poor generalization to unseen motion patterns. Second, existing methods primarily introduce simple visibility penalties during training, while lacking predictive compensation or recovery mechanisms under visual loss. As task difficulty increases and more aggressive maneuvers become necessary, intermittent perception loss shifts the policy inputs beyond the training distribution, thereby compromising the reliability of end-to-end control.

\begin{figure}[t]
    \centering
    \includegraphics[width=\columnwidth]{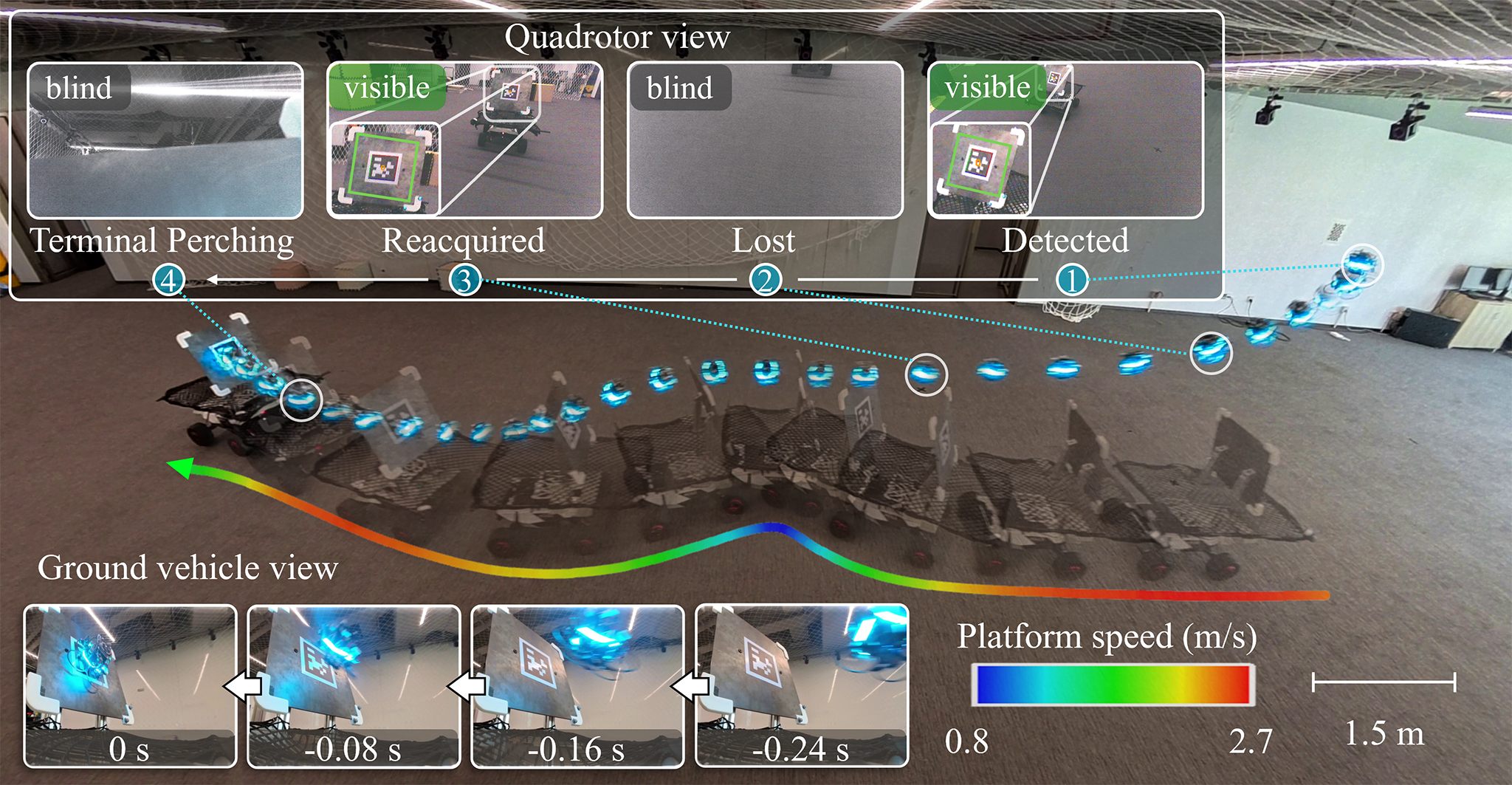} 
    \caption{Real-world demonstration of PerchRL. The ground vehicle equipped with a $70^\circ$ inclined platform tracks a randomly generated spline trajectory with periodically varying forward speed. The top and bottom inset images show additional views of the aerial perching process from cameras mounted on the quadrotor and ground vehicle, respectively.}
    \label{fig:hard1_experiment}
    \vspace{-0.4cm}
\end{figure}

We identify the core challenge as maintaining reliable control under intermittent visual loss, which requires simultaneously addressing three coupled difficulties: (a) the platform state estimate is interrupted during visual loss, causing abrupt input distribution shifts; (b) even when a predictive estimate is available to bridge the gap, the policy cannot gauge its reliability and may continue to rely on an estimate that has significantly diverged; and (c) under underactuated dynamics, the maneuvers needed to re-establish visibility may conflict with those required for timely approach and perching. Failure to address any single one leads to task failure.

To address these challenges, we propose PerchRL, a vision-based agile perching framework with zero-shot sim-to-real transfer capability. PerchRL adopts a two-stage learning strategy consisting of state-based pre-training followed by vision-based fine-tuning. In the first stage, we employ dynamics-aware spatiotemporal trajectory randomization to generate diverse, smooth, and dynamically feasible target motions, preventing overfitting to specific motion patterns. Temporal encoding of historical observations further assists the policy in capturing latent motion trends, improving generalization to unseen trajectories. In the second stage, we propose a visibility-aware hybrid guidance framework. During visual loss, an estimator maintains a continuously evolving platform state estimate, preventing abrupt input discontinuity. To make potential divergence of the predictive estimate perceptible to the policy, we augment the observation with a reliability indicator, informing the policy when the estimate should no longer be fully trusted. Complementing this implicit augmentation, active perception rewards explicitly incentivize visibility re-establishment. Together, this hybrid design enables the policy to sustain aggressive approach maneuvers while the estimate remains reliable, deferring recovery until confidence degrades, flexibly balancing perching progress with perception maintenance.

Overall, PerchRL presents, to the best of our knowledge, the first RL-based framework to demonstrate vision-based agile attachment to a rapidly moving platform with a large fixed inclination, while explicitly addressing intermittent loss of visual target-pose measurements. The main contributions are summarized as follows:

\begin{enumerate}
    \item We formalize the challenge of maintaining agile and reliable control under intermittent visual loss, identifying that visual loss disrupts state estimation, erodes the trustworthiness of predictive estimates, and forces a conflict between visibility recovery and timely perching.
    \item We introduce a dynamics-aware spatiotemporal trajectory randomization method that generates diverse, smooth, and dynamically feasible platform motions, preventing overfitting and improving generalization to unseen motion patterns.
    \item We propose a visibility-aware hybrid guidance framework that enables the policy to sustain aggressive approach maneuvers while the estimate remains reliable and defer recovery until confidence degrades, flexibly balancing perching progress with perception maintenance.
    \item We conduct extensive simulation and real-world experiments, demonstrating the feasibility, stability, and real-time performance of PerchRL.
\end{enumerate}

\section{Related Work}
\label{sec:related_work}

\subsection{Model-Based Methods for Aerial Perching}
\label{subsec:model_based_aggressive_perching}

In the past, model-based approaches have been widely employed for aerial perching. Early works primarily focused on stationary inclined platforms \cite{mellinger2012trajectory, mao2021aggressive}, while more recent studies \cite{ji2022real, liu2025quadrotors} have extended to moving inclined platforms by explicitly modeling platform dynamics. To prevent visual perception loss, various strategies have been proposed to maintain stable visibility. Mao et al. \cite{mao2023robust} proposed a generalizable FOV constraint with a lightweight representation, and Gao et al. \cite{gao2023adaptive} incorporated perception constraints based on the pinhole camera model into the nonlinear optimization framework of \cite{ji2022real} to keep the platform within the camera's FOV. However, these methods rely on accurate motion prediction and consume substantial computational resources for visibility analysis and optimization during trajectory generation. As a result, they often struggle under fast and unpredictable platform motion, where high-frequency replanning is essential. 

\subsection{RL-Based Methods for Aerial Perching}
\label{subsec:learning_based_aggressive_perching}

Recently, RL-based methods have emerged as a promising solution for aerial perching. Early work \cite{polvara2018toward} employed a hierarchical Deep Q-Networks (DQNs) for autonomous perching on a static landing pad. Later studies further extended aerial perching to inclined \cite{kooi2021inclined,van2024eagerx} or dynamic \cite{rodriguez2019deep,goldschmid2024reinforcement} platforms using more advanced RL algorithms and simulators. However, most existing approaches are still restricted to fixed or highly structured platform motion, such as linear or circular trajectories, and exhibit poor generalization to unseen tasks. Moreover, achieving robust vision-based perching under intermittent visual perception loss remains highly challenging. Existing methods typically introduce active-perception rewards during training to encourage policies to maintain stable visibility. Wang et al. \cite{wang2023vision} employed a reward proportional to the platform’s pixel occupancy ratio, while Ladosz et al. \cite{ladosz2024autonomous} introduced a simple constant penalty when the platform leaves the quadrotor’s FOV. However, the lack of predictive compensation or recovery mechanisms limits their robustness under visual perception loss. Specifically, these methods typically assume that the learned policy maintains reliable visual perception throughout the entire perching task, without explicitly accounting for perception loss. As task difficulty increases and more aggressive flight maneuvers become necessary, intermittent perception loss may shift the policy inputs beyond the training distribution, thereby compromising the reliability of end-to-end control.


\section{Methodology}
\label{sec:methodology}

\subsection{Algorithm Overview}
\label{subsec:algorithm_overview}

\begin{figure}[t]
    \centering
    \includegraphics[width=\columnwidth]{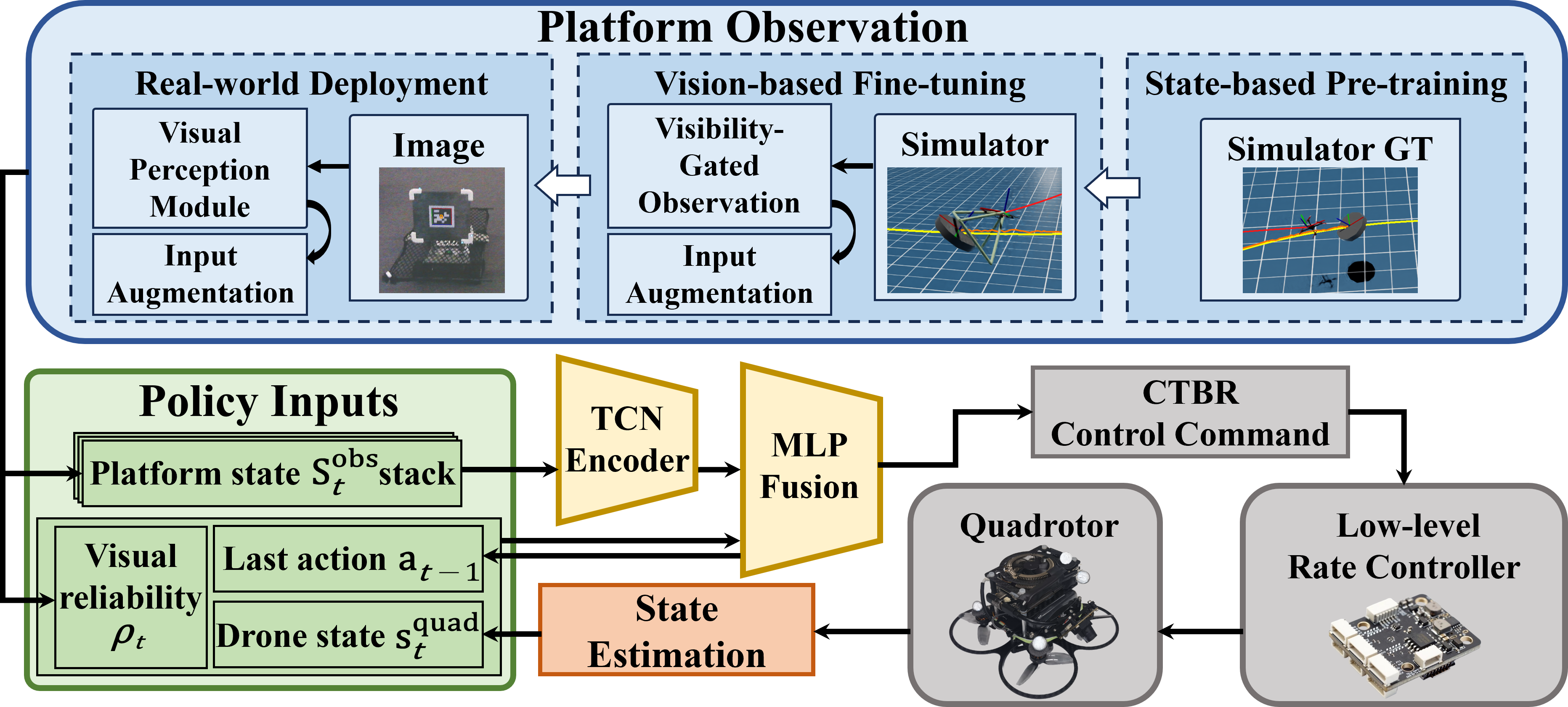}
    \caption{System overview of PerchRL, including the training pipeline and deployment architecture.}
    \label{fig:overview}
    \vspace{-0.6cm}
\end{figure}

Fig.~\ref{fig:overview} provides a high-level overview of PerchRL. It adopts a two-stage learning strategy that first learns agile interception and terminal alignment under diverse platform motions, then adapts this behavior to intermittent visual feedback, thereby reducing training difficulty. The policy acquires platform observations in different forms across stages. First, during state-based pre-training, the policy directly accesses ground-truth platform states from the simulator. Next, during vision-based fine-tuning, it receives noisy visual observations from the visibility-gated observation model proposed in Section~\ref{subsec:Control-Aware Observation Modeling}. Finally, during real-world deployment, a plug-and-play visual detection module is employed to provide real-time visual observations from onboard camera images. Details of the training pipeline and algorithmic design are provided in Sections~\ref{subsec:vision free perching} and~\ref{subsec:vision based perching}. 


\begin{figure*}[t]
    \centering
    \includegraphics[width=17.5cm]{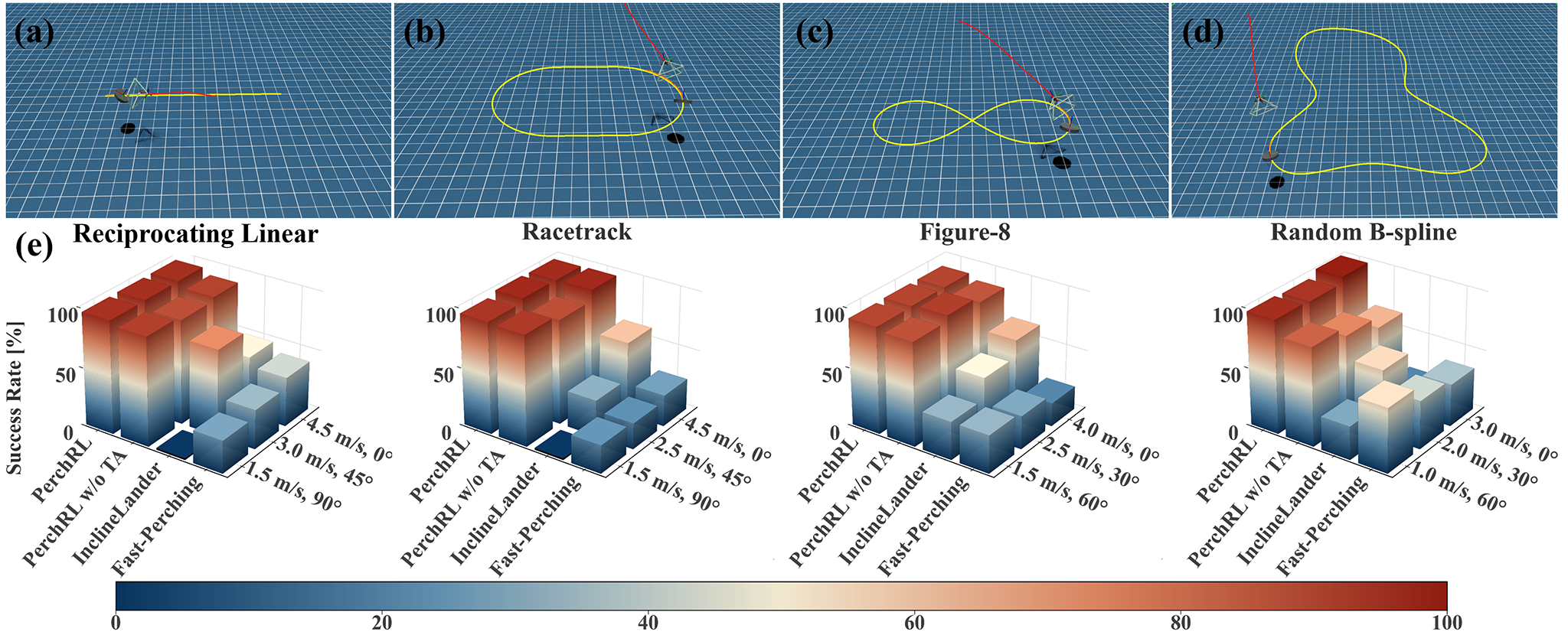}
    \caption{Baseline and ablation studies of state-based perching under diverse platform motion patterns. (a)-(d) Visualization of evaluated trajectories in the simulator. (e) Success rate comparison across different methods and scenarios.}
    \label{fig:benchmark}
    \vspace{-0.5cm}
\end{figure*}

\subsection{State-Based Policy Pre-Training}
\label{subsec:vision free perching}

The state-based perching task is formulated as a Markov decision process (MDP), defined as a tuple $\mathcal{M} = (\mathcal{S}, \mathcal{A}, \mathcal{P}, \mathcal{R}, \gamma)$. The goal is to learn an optimal policy $\pi^{*}$ that maximizes the expected cumulative reward: $\pi^* = \arg \max_{\pi} \mathbb{E} [ \sum_{t=0}^T \gamma^t R(\mathbf{s}_t, \mathbf{a}_t) ]$. 


\textit{\textbf{Platform Motion:}} The moving platform is modeled as a standard differential-drive robot governed by nonholonomic planar kinematics. To prevent the policy from overfitting to specific motion patterns, we generate a randomized closed-loop reference trajectory at the beginning of each episode. The trajectory is represented as a periodic uniform cubic B-spline $p(t)$ to exploit its smoothness, local controllability, and analytically tractable derivatives for generating dynamically consistent motions. $p(t)$ is defined by $M$ control points $\mathcal{C}=\{\mathbf{C}_0,\mathbf{C}_1,\ldots,\mathbf{C}_{M-1}\}$, which are constructed in polar coordinates as follows:
\begin{equation}
\begin{aligned}
\theta_i &= \frac{2\pi i}{M},                 &
\tilde{r}_i &\sim \mathcal{U}(r_{\min},r_{\max}),\\
r_i &= \mathcal{F}_{\mathrm{LPF}}(\tilde{r}_i), &
\mathbf{C}_i &=
[r_i\cos\theta_i,\;r_i\sin\theta_i,\;c_z]^\top
\end{aligned}
\label{eq:control_point_sampling}
\end{equation}
where $r_{\min}$ and $r_{\max}$ define the radial sampling range, and $c_z$ denotes the nominal platform height. The circular low-pass filter $\mathcal{F}_{\mathrm{LPF}}(\cdot)$ suppresses abrupt radial changes along the angular direction, providing initial smoothing.

Furthermore, we account for the motion constraints of real differential-drive robots and propose a global curvature checking and optimization procedure. Specifically, we first perform a global curvature check of the trajectory and determine its maximum curvature $\kappa_{\max}$, which must satisfy the following constraint:
\begin{equation}
\kappa_{\max}
\leq
\frac{\dot{\psi}_{\lim}}
{\max(v_{\mathrm{des}},\epsilon)}
\label{eq:curvature_constraint}
\end{equation}
where $\dot{\psi}_{\lim}$ denotes the maximum yaw rate, $v_{\mathrm{des}}$ is the nominal traversal speed, and $\epsilon$ is a small positive introduced for numerical stability. If the constraint is violated, the radial deviations are uniformly contracted as:
\begin{equation}
r_i'=\bar{r}+\alpha_{\mathrm{bin}}(r_i-\bar{r}),
\qquad \forall i=0,1,\cdots,M-1
\label{eq:radial_contraction}
\end{equation}
where $r_i$ and $r_i'$ denote the original and updated radii of the $i$-th control point, respectively, and $\bar r$ is the mean radius of all control points. $\alpha_{\mathrm{bin}}$ is initialized to 1 and updated via a binary search strategy until Eq.~\eqref{eq:curvature_constraint} is satisfied. This procedure enforces the curvature constraint while preserving as much geometric diversity as possible.




In addition to spatial randomization, we introduce temporally correlated speed perturbations based on the following discretized Ornstein--Uhlenbeck process:
\begin{equation}
\begin{aligned}
\beta_{\mathrm{ou}} &= \frac{\Delta t}{\tau+\Delta t}\\
\eta_t &= (1-\beta_{\mathrm{ou}})\eta_{t-1}
+\sigma\sqrt{\beta_{\mathrm{ou}}}\varepsilon_t\\
v_t^{\mathrm{pert}} &= v_{\mathrm{des}}\cdot\exp(\eta_t)
\end{aligned}
\label{eq:velocity_perturbation}
\end{equation}
where $\Delta t$ denotes the simulation time step, $\varepsilon_t$ is standard Gaussian noise, and $\tau$ and $\sigma$ control the temporal smoothness and perturbation magnitude, respectively. This temporally correlated perturbation better captures the smooth velocity evolution of real platforms while avoiding physically implausible speed fluctuations between consecutive frames caused by independent random sampling.



Overall, the proposed dynamics-aware spatiotemporal randomization method improves the consistency between the reference motion and the motion characteristics of real differential-drive robots. It also exposes the policy to diverse, smooth, and dynamically feasible platform motions, thereby enhancing its robustness and generalization capability.

\textit{\textbf{Observations and Actions:}} The policy observation is defined as $\mathbf{s}_t = [\mathbf{s}_t^{\text{quad}}, \mathbf{S}_t^{\text{obs}}, \rho_{t}, \mathbf{a}_{t-1}]^\top$, where $\mathbf{s}_t^{\text{quad}}=[{\mathbf{v},\mathrm{vec}(\mathbf{R}(\mathbf{q}))}]^\top$ denotes the drone’s velocity and the vectorized rotation matrix in the world frame. $\mathbf{S}_t^{\text{obs}}=[\mathbf{s}_{t-H+1}^{\text{obs}}, \dots, \mathbf{s}_t^{\text{obs}}]$ is a history of $H$ platform observations, with each $\mathbf{s}_{i}^{\text{obs}} =[{\mathbf{p}_i^{\mathrm{rel}}, \mathbf{z}_i, \mathbf{V}_i}, t_i^{\mathrm{rel}}]^\top$ comprising the platform relative position, normal vector, velocity $\mathbf{V}_i = [v_i^x,v_i^y,\omega_i]^\top$, as well as the relative timestamp. $\rho_{t}$ encodes visual reliability (see Section \ref{subsec:compact-belief-trust-aware-prediction}). During this stage, it is fixed to $1$ for input consistency. $\mathbf{a}_{t-1}$ denotes the previous action.

To enable agile maneuvers, we adopt mass-normalized collective thrust and body rates (CTBR), which improves robustness to the sim-to-real gap~\cite{kaufmann2022benchmark}. The policy generates normalized actions $\mathbf{a}_t = [\tilde{\boldsymbol{\omega}}_{\text{cmd}}, \tilde{T}_{\text{cmd}}] \in \mathbb{R}^{4}$, which are mapped to actual control commands as follows:
\begin{equation}
\begin{aligned}
T_{\text{cmd}} &= \frac{1 - \tilde{T}_{\text{cmd}}}{2} T_{\text{min}} + \frac{1 + \tilde{T}_{\text{cmd}}}{2} T_{\text{max}} \\ \boldsymbol{\omega}_{\text{cmd}} &= \omega_{\text{max}}\tilde{\boldsymbol{\omega}}_{\text{cmd}} 
\end{aligned}
\end{equation}
where $T_{\text{min}}, T_{\text{max}}, \omega_{\text{max}}$ denote the thrust bounds and maximum body-rate limits, respectively.

\textit{\textbf{Policy Architecture:}} For temporal augmentation, we employ a temporal convolutional network (TCN) with three 1D convolutional layers to encode $\mathbf{S}_t^{\text{obs}}$, capturing latent temporal representations of platform motion. The temporal features are concatenated with other instantaneous inputs and fed into an MLP head to output the actions. In addition, an asymmetric actor-critic architecture is introduced to further augment temporal modeling. Specifically, we augment the critic inputs with future platform states sampled from its reference trajectory. The privileged input consists of a sequence of future observations $\mathbf{S}_t^{\text{priv}}=[\mathbf{s}_{t+1}^{\text{obs}}, \dots, \mathbf{s}_{t+N}^{\text{obs}}]^\top$ with $N=15$, sharing the same structure as $\mathbf{S}_t^{\text{obs}}$. Leveraging both historical observations and privileged future information, this design leads to more accurate and comprehensive value estimation, thereby enhancing the effectiveness of actor training.


\begin{figure}[t]
    \centering
    \includegraphics[width=\columnwidth]{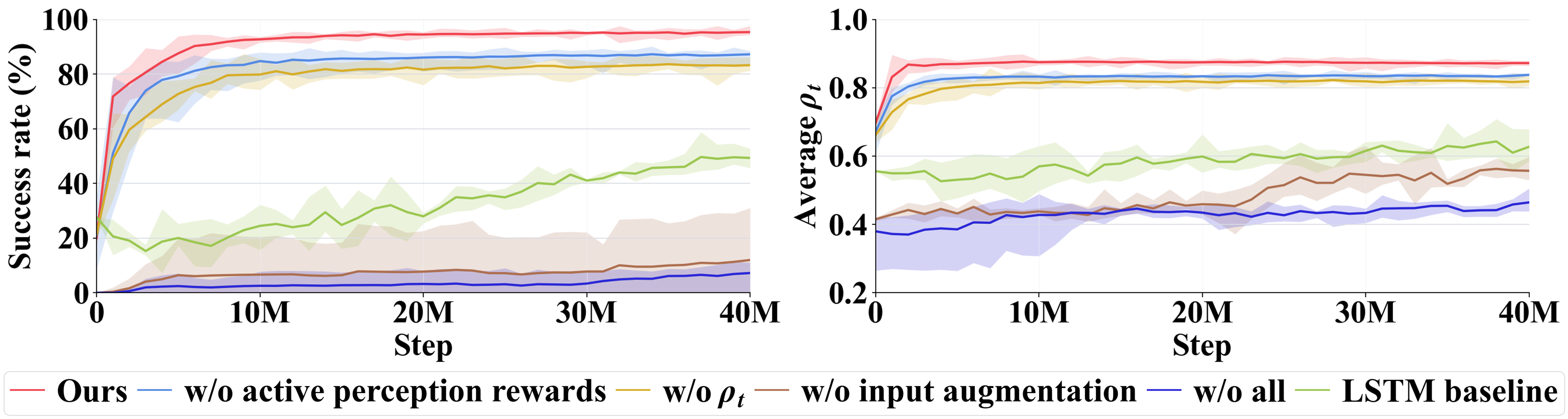}
    \caption{The training curves of success rate and average $\rho_t$ for our method and baseline or ablated variants on platforms with up to $2.0\,\mathrm{m/s}$ velocity and $45^\circ$ inclination. Here, the average $\rho_t$ reflects visual perception stability during perching. For methods lacking full input augmentation, an equivalent $\rho_t$ is computed only for evaluation and is inaccessible to the policy.}
    \label{fig:ablation}
    \vspace{-0.4cm}
\end{figure}

\textit{\textbf{Reward Function:}}
The immediate reward combines a dense shaping reward with a sparse terminal reward:
\begin{equation}
r_t = r_t^{\mathrm{shaping}} + r_t^{\mathrm{terminal}}
\end{equation}
On the one hand, the shaping reward $r_t^{\text{shaping}}$ provides dense intermediate feedback to efficiently guide the policy toward agile and dynamically feasible perching maneuvers. On the other hand, the terminal reward $r_t^{\text{terminal}}$ is designed to encourage successful perching, with a uniform reward for all successful cases and an accuracy-dependent bonus for more precise terminal state alignment.  

The dense shaping reward is decomposed as:
\begin{equation}
r_t^{\mathrm{shaping}}
=r_t^{\mathrm{guide}}
+r_t^{\mathrm{align}}
+r_t^{\mathrm{act}}
+r_t^{\mathrm{aggr}}
+r_t^{\mathrm{time}}
\end{equation}
where $r_t^{\mathrm{guide}}$,
$r_t^{\mathrm{act}}$, $r_t^{\mathrm{aggr}}$, and
$r_t^{\mathrm{time}}$ follow the design principles of \cite{wu2025whole}. Specifically, $r_t^{\mathrm{guide}}$ guides the quadrotor toward the
platform by reducing its relative pose error, $r_t^{\mathrm{act}}$ promotes small and temporally smooth control actions, $r_t^{\mathrm{aggr}}$ penalizes violations of the prescribed motion limits, and $r_t^{\mathrm{time}}$ discourages unnecessarily long maneuvers. In addition, $r_t^{\mathrm{align}}$ is introduced to further enhance state alignment during the terminal perching phase:
\begin{equation}
\begin{gathered}
\phi_1(x)=\bigl((1+\cos x)/2\bigr)^2,\qquad
\phi_2(x)=(1+x^2)^{-1},\\[-1mm]
r_t^{\mathrm{align}}
=r_t^{\mathrm{guide}}\!\left[
k_\theta\cdot\phi_1(\theta_t^{\mathrm{rel}})
+\phi_2\!\left(k_v\cdot\lVert\mathbf v_t^{\mathrm{rel}}\rVert_2\right)
+\phi_2(\dot{\psi}_t)
\right].
\end{gathered}
\end{equation}
where $\theta_t^{\mathrm{rel}}$,
$\mathbf v_t^{\mathrm{rel}}$, and $\dot{\psi}_t$ denote the
normal-alignment error, relative velocity, and body-frame yaw rate, respectively, while $k_\theta$ and $k_v$ are scale factors.

The sparse terminal reward $r_t^{\text{terminal}}$ is employed to explicitly enforce terminal constraints and improves perching accuracy. At each timestep, the episode is classified as \textbf{Ongoing}, \textbf{Failure}, or \textbf{Success} through a hierarchical evaluation procedure. We first check for violation conditions, including crashing, exceeding the predefined workspace, and reaching the maximum episode length, any of which immediately triggers failure termination. Otherwise, contact is detected using the criterion proposed in \cite{ji2022real}. If no contact is detected, the episode remains ongoing. Once contact occurs, the drone state is transformed into the platform coordinate frame to obtain the relative state $\mathbf{s}_t^{\text{rel}}=[\Delta d_t,\Delta v_t,\Delta\theta_t]^\top$, where $\Delta d_t$, $\Delta v_t$, and $\Delta\theta_t\in\mathbb{R}^+$ denote the relative tangential distance, relative tangential velocity, and normal angular difference, respectively. Given tolerance thresholds $\boldsymbol{\delta}_{\text{tol}}=[\delta_d,\delta_v,\delta_\theta]^\top$, successful perching is achieved if and only if $\mathbf{s}_t^{\text{rel}}\leq\boldsymbol{\delta}_{\text{tol}}$ element-wise. Otherwise, the episode terminates with failure. Accordingly, the episode status is defined as:
\begin{equation}
c_t =
\begin{cases}
\mathrm{Fail}, & \text{if a violation occurs},\\
\mathrm{Ongoing}, & \text{if no contact is detected},\\
\mathrm{Success}, & \text{if } \mathbf{s}_t^{\mathrm{rel}}
\leq \boldsymbol{\delta}_{\mathrm{tol}},\\
\mathrm{Fail}, & \text{otherwise},
\end{cases}
\label{eq:episode_status}
\end{equation}
and the corresponding terminal reward is:
\begin{equation}
r_t^{\mathrm{terminal}} =
\begin{cases}
\lambda_s(1+w_s), & \text{if } c_t=\mathrm{Success},\\
\lambda_f, & \text{if } c_t=\mathrm{Fail},\\
0, & \text{if } c_t=\mathrm{Ongoing}.
\end{cases}
\label{eq:terminal_reward}
\end{equation}
where $\lambda_s,\lambda_f$ are adjustable weights, and the accuracy-dependent bonus is computed as:
\begin{equation}
\begin{aligned}
f_l(s,\delta) &= ((s-\delta)/\delta)^2 \\
w_s &=
\operatorname{mean}\left(
\mathbf{f}_l(
\mathbf{s}_t^{\text{rel}},
\boldsymbol{\delta}_{\text{tol}})
\right)
\end{aligned}
\end{equation}
The success reward $r_t^{\text{terminal}}\in[\lambda_s,2\lambda_s]$ provides a consistent reward for satisfying the success condition while assigning a larger bonus to states with higher alignment accuracy. This encourages the policy to achieve more precise and stable perching beyond merely satisfying the terminal constraints.

\subsection{Vision-Based Policy Fine-Tuning}
\label{subsec:vision based perching}

The vision-based perching task is formulated as a partially observable Markov decision process (POMDP), defined as $\mathcal{M} = (\mathcal{S}, \mathcal{A}, \mathcal{O}, \mathcal{P}, \mathcal{R}, \Omega, \gamma)$. The platform state is only partially observable from visual observations due to visibility constraints. To successfully fine-tune the policy trained in Section \ref{subsec:vision free perching} to acquire vision-based perching capability, it is necessary to establish an appropriate visual perception simulation pipeline along with more effective training strategies.

\subsubsection{Visibility-Gated Observation Model}  
\label{subsec:Control-Aware Observation Modeling}

Existing vision-based perching policies typically adopt end-to-end learning directly from raw images, requiring extensive domain randomization and remaining sensitive to appearance variations and the sim-to-real gap. In contrast, we adopt a modular visual observation formulation to focus on agile control while avoiding dependence on a specific visual detector.


Specifically, we assume an upstream visual detector that provides intermittent 6-DoF pose estimates of the platform in the world frame. Visual measurements are subject to a visibility-gated observation process. At each time step, the simulator first checks whether the elapsed time since the last visual update exceeds the detector period $1/f_{\mathrm{det}}$, where $f_{\mathrm{det}}$ denotes the predefined detection frequency. A new measurement is triggered only if this condition is satisfied. When a measurement is triggered, a simulated downward-facing pinhole camera model is used to evaluate visibility based on the current states of the drone and platform. If the platform is not visible, the measurement is discarded. Otherwise, a visual observation noise model is applied to transform the ground-truth pose $\mathbf{T}_{t}^{p}$ into a noisy observation $\hat{\mathbf{T}}_{t}^{p}$, accounting for onboard perception degradations and improving the policy's robustness to real-world uncertainty. The noise model is adapted and improved from the approach proposed in \cite{gao2023adaptive}, and exhibits anisotropic, geometry-dependent characteristics together with systematic biases.

Here, ``vision-based'' refers to system-level sensing, as platform observations originate from onboard camera images during deployment, while the policy consumes intermittent 6-DoF pose estimates from the upstream detector instead of raw pixels.


\subsubsection{Visibility-Aware Policy Learning with Hybrid Guidance}
\label{subsec:compact-belief-trust-aware-prediction} 

The platform becomes fully unobservable once it leaves the FOV, requiring effective and robust perception recovery. Although prior works~\cite{wu2025whole, shin2026vision} have demonstrated the potential of temporal neural networks with historical observations for mitigating the impact of intermittent observation loss in other domains, we find that directly applying this paradigm is insufficient for robust vision-based perching under intermittent visual loss (see Section \ref{subsec:baseline_and_ablation}). To address this challenge, we propose a hybrid training scheme that explicitly encourages visibility maintenance via active-perception rewards and implicitly enhances visibility-aware behavior through an input augmentation mechanism.

\textit{\textbf{Visibility-Aware Input Augmentation:}} The proposed input augmentation mechanism provides the policy with two types of information, including smoothed platform motion features and the duration of the current visual perception loss. Specifically, we construct a state estimation module based on an extended Kalman filter (EKF) driven by the constant turn rate and velocity (CTRV) model \cite{huang2022survey}. At each step, the filter performs state prediction, while measurement updates gated by the Mahalanobis distance \cite{chang2014robust} are applied only when visual observations $\hat{\mathbf{T}}_{t}^{p}$ are available. We derive $\hat{\mathbf{s}}_t^{\text{obs}}$ from the internal filter state as an estimate of $\mathbf{s}_t^{\text{obs}}$ defined in Section \ref{subsec:vision free perching}. Moreover, although $\hat{\mathbf{s}}_t^{\text{obs}}$ preserves observation continuity during visual loss, this input design does not encode the occurrence or duration of visual loss. To address this issue, the module maintains an indicator $\rho_t$ representing the current visual perception reliability as follows:
\begin{equation}
\begin{aligned}
\rho_t = \begin{cases} 
\exp\left(-\lambda_{\rho} \cdot T_t^{\mathrm{loss}}  \right), & \text{during visual loss,} \\ 
1, & \text{otherwise.} 
\end{cases}
\end{aligned}
\end{equation}
where $\lambda_{\rho}$ is a scale factor and $T_t^{\mathrm{loss}}$ denotes the duration of the current visual loss. Based on the definition of $\rho_t$, a decreasing $\rho_t$ indicates a decline in the reliability of $\hat{\mathbf{s}}_t^{\text{obs}}$ due to prolonged visual loss, which informs the policy to actively adjust its sensing behavior to re-establish reliable visual perception. The proposed method augments the raw visual observations with visibility-aware information, implicitly enhancing the policy’s ability to adaptively balance perching maneuvers and visibility maintenance under intermittent visual loss. 


\textit{\textbf{Active Perception Rewards:}} The augmented inputs expose the information of visibility-aware control, but do not specify how to balance perception recovery against task progress. We therefore introduce an auxiliary reward $r_t^{\mathrm{active}}$ that penalizes estimation error and encourages camera--target alignment:
\begin{subequations}
\begin{align}
g_t^{\mathrm{loss}} &= 1+\alpha \cdot \mathrm{clip}\left(T_t^{\mathrm{loss} } / \tau \,,0\,,1 \right) \\
r_t^{\mathrm{est}}&=-\beta \cdot  g_t^{\mathrm{loss}} \cdot \left\|\hat{\mathbf{p}}_{t}-\mathbf{p}_{t}\right\| \\
r_t^{\mathrm{perc}}&=-\eta \cdot \left(1-\mathbf{v}_{\mathrm{cam}}^{\top}\mathbf{d}_{\mathrm{target}}\right) \\
r_t^{\mathrm{active}}&=r_t^{\mathrm{est}}+r_t^{\mathrm{perc}}
\end{align}
\end{subequations}
where $\alpha,\beta,\eta,\tau$ are tunable hyperparameters. $r_t^{\mathrm{est}}$ penalizes actions that increase estimation error $\left\|\hat{\mathbf{p}}_{t}-\mathbf{p}_{t}\right\|$ or lead to prolonged perception loss. $r_t^{\mathrm{perc}}$ promotes perception awareness by encouraging alignment between the camera optical axis $\mathbf{v}_{\mathrm{cam}}$ and target direction $\mathbf{d}_{\mathrm{target}}$. The final immediate reward is defined as $r_t=r_t^{\text{shaping}} + r_t^{\text{terminal}}+r_t^{\text{active}}$. Overall, input augmentation supports decision-making under missing feedback, while explicit reward guidance facilitates learning the corresponding control behavior.



\begin{table}[t]
\centering
\caption{Initialization distributions for training and evaluation}
\label{tab:initial_reset_conditions}
\scriptsize
\setlength{\tabcolsep}{3pt}
\renewcommand{\arraystretch}{1.12}

\begin{tabularx}{\columnwidth}{@{}X c@{}}
\toprule
\textbf{Initialization variable} & \textbf{Distribution} \\
\midrule

\multicolumn{2}{@{}l}{\textbf{Platform motion and configuration}}\\[-1pt]
\hspace{0.8em}Desired speed $v_{\mathrm{des}}^{p}$ [m/s]
    & $\mathcal{U}(0,\,5.0)$ \\
\hspace{0.8em}Constant pitch angle $\theta_c^{p}$ [$^\circ$]
    & $\mathcal{U}(0,\,90)$ \\

\addlinespace[3pt]
\multicolumn{2}{@{}l}{\textbf{Quadrotor initial state}}\\[-1pt]
\hspace{0.8em}Relative altitude $\Delta z_0$ [m]
    & $\mathcal{U}(1.2,\,2.0)$ \\
\hspace{0.8em}Relative horizontal distance $\Delta d_0$ [m]
    & $\mathcal{U}(1.2,\,3.5)$ \\
\hspace{0.8em}Relative azimuth offset $\Delta\gamma_0$ [$^\circ$]
    & $\mathcal{U}(-60,\,60)$ \\
\hspace{0.8em}Initial roll/pitch $(\phi_0^q,\theta_0^q)$ [rad]
    & $\mathcal{U}(-0.6,\,0.6)$ \\
\hspace{0.8em}Initial horizontal velocity $(v_{0,x}^q,v_{0,y}^q)$ [m/s]
    & $\mathcal{U}(-0.1,\,0.1)$ \\
\hspace{0.8em}Initial vertical velocity $v_{0,z}^q$ [m/s]
    & $\mathcal{U}(-0.2,\,0.2)$ \\
\hspace{0.8em}Initial roll/pitch rate
$(\dot{\phi}_0^q,\dot{\theta}_0^q)$ [rad/s]
    & $\mathcal{U}(-0.2,\,0.2)$ \\
\hspace{0.8em}Initial yaw rate $\dot{\psi}_0^q$ [rad/s]
    & $\mathcal{U}(-0.1,\,0.1)$ \\

\bottomrule
\end{tabularx}
\vspace{-12pt} 
\end{table}
\section{Experiments}
\label{sec:experiments}

\subsection{Simulation and Training}
\label{subsec:simulation_and_training}

The training of PerchRL is performed using proximal policy optimization (PPO)~\cite{schulman2017proximal}, and the toolchain is developed based on the open-source Omnidrones~\cite{xu2024omnidrones}. To accelerate the training process, we employ large-scale parallel simulation with 8192 environments running on a workstation equipped with an Intel i7-14700KF CPU and an NVIDIA RTX 4080 GPU. The simulator operates at 100 Hz with a timestep of 0.01 s, and the maximum episode length is set to 500 steps.

At the start of each episode, the platform’s motion and configuration parameters are randomly sampled, including its desired speed and inclination angle. The quadrotor is subsequently initialized at a random position within a sector-shaped region behind and above the platform, while its initial attitude and velocity are randomly sampled. In particular, during vision-based policy training, we resample the initial configuration until the platform is visible to the quadrotor initially. The initialization parameters used during both training and evaluation are summarized in Table \ref{tab:initial_reset_conditions}.

In addition, the following domain randomization strategies are applied to mitigate the sim-to-real gap, thereby improving the robustness and generalization of the trained policy:

\begin{itemize}{\leftmargin=1em}
    \setlength{\topmargin}{0pt}
    \setlength{\itemsep}{0pt}
    \setlength{\parsep}{0pt}
    \setlength{\parskip}{0pt}
    \item To account for non-linearities and environmental perturbations, three key dynamic parameters, namely mass, inertia, and thrust-to-weight ratio, are independently randomized within $\pm30\,\%,\pm10\,\%,\pm15\,\%$ of their nominal values, respectively.
    \item To account for mounting misalignments and calibration errors, the camera tilt angle is uniformly sampled within $\pm5^\circ$ of its nominal value.
    \item To account for variations in perception update rates and visual processing latency, the detector update frequency $f_{\mathrm{det}}$ described in Section~\ref{subsec:Control-Aware Observation Modeling} is uniformly randomized within $[20,\,60]$ Hz.
\end{itemize}

\subsection{Baseline and Ablation Studies}
\label{subsec:baseline_and_ablation}

\textit{\textbf{State-based Perching:}} We first evaluate state-based perching against two baselines: a model-based method Fast-Perching \cite{ji2022real} and an RL-based method InclineLander \cite{kooi2021inclined}. Additionally, we introduce an ablated variant of PerchRL without temporal augmentation, which directly maps instantaneous observations to actions through an MLP head, referred to as PerchRL w/o TA.

We further introduce three fixed and structured trajectories for comprehensive evaluation, which are the Reciprocating Linear, Racetrack, and Figure-8 trajectories, respectively, as shown in Fig.~\ref{fig:benchmark}(a)-(c). For fair comparison, all RL-based baselines are trained with dedicated policies for each trajectory for up to 40M steps to ensure convergence, and evaluated across three scenarios with varying platform velocities and inclinations.

Fig.~\ref{fig:benchmark} reports success rates across all methods and scenarios. PerchRL and its variant consistently outperform other baselines, highlighting the limitations of the latter. Fast-Perching predicts platform motion under a constant-velocity assumption, leading to degraded performance under complex platform motion. Meanwhile, InclineLander shows limited performance under challenging conditions, relying on sparse terminal rewards without dense intermediate shaping. In addition, the ablated variant achieves comparable performance on the additionally introduced trajectories, but suffers a training bottleneck on B-spline trajectories and exhibits limited generalization. Performance further degrades with increasing platform velocity, indicating limited adaptability to rapid and irregular platform motion. In contrast, with the temporal augmentation methods proposed in Section \ref{subsec:vision free perching}, our method demonstrates strong generalization and achieves the highest success rate across all scenarios.




\begin{figure}[t]
    \centering
    \includegraphics[width=\columnwidth]{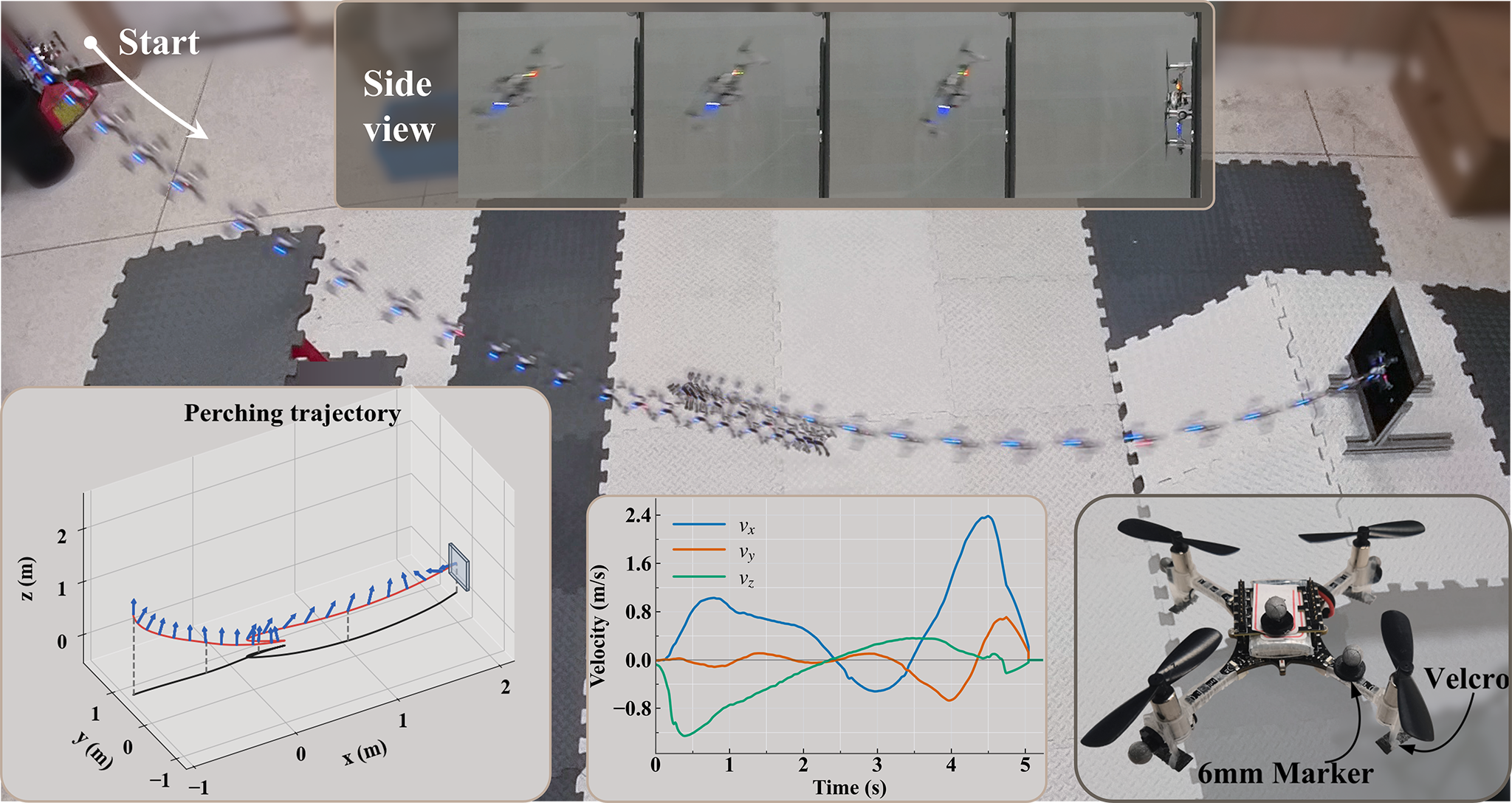}
    \caption{Real-world demonstration of PerchRL for state-based perching on a $90^\circ$ inclined surface with a lightweight Crazyflie. The bottom inset images show the Crazyflie used in the experiments, its velocity profile, and a visualization of the 3D perching trajectory, where the blue arrow denotes the body z-axis of the drone.}
    \label{fig:crazyflie}
    \vspace{-0.6cm}
\end{figure}

\textit{\textbf{Vision-based Perching:}} Due to the difficulty of the studied setting, prior end-to-end vision-based methods do not provide directly comparable baselines. We therefore implement a task-matched LSTM recurrent baseline mapping intermittent 6-DoF detections and ego-related inputs to CTBR commands, trained with the same two-stage training strategy. We further conduct ablation studies to evaluate the contributions of the input augmentation method and active perception rewards proposed in Section \ref{subsec:compact-belief-trust-aware-prediction} to perching performance. Specifically, we introduce four ablated variants of PerchRL and train all methods with the same pretrained model and environment settings for fair comparison. For methods without input augmentation, instantaneous velocity is estimated from raw visual observations via temporal differencing, with $\rho_t$ set to 1 to preserve the input structure of the policy network.


As illustrated in Fig.~\ref{fig:ablation}, the LSTM-Baseline achieves only moderate performance, highlighting the difficulty of learning robust perching under intermittent visual loss. The ablated variant without any auxiliary mechanisms struggles with effective exploration, resulting in the poorest performance. Moreover, the proposed input augmentation strategy contributes substantially to vision-based training, while removing it leads to a significant performance drop. Notably, independently utilizing the augmentation strategy is sufficient to achieve competitive performance. In contrast, although active perception rewards improve training efficiency and further enhance final performance, they are insufficient for learning robust perching under intermittent visual loss. In addition, removing $\rho_t$ eliminates the policy’s explicit awareness of visual perception loss, forcing it to fully trust and rely on $\hat{\mathbf{s}}_t^{\text{obs}}$ even when it has significantly deviated from the ground truth, leading to degraded performance. Overall, PerchRL achieves the best performance and visual perception stability while also exhibiting faster training convergence.

\begin{figure}[t]
    \centering
    \includegraphics[width=\columnwidth]{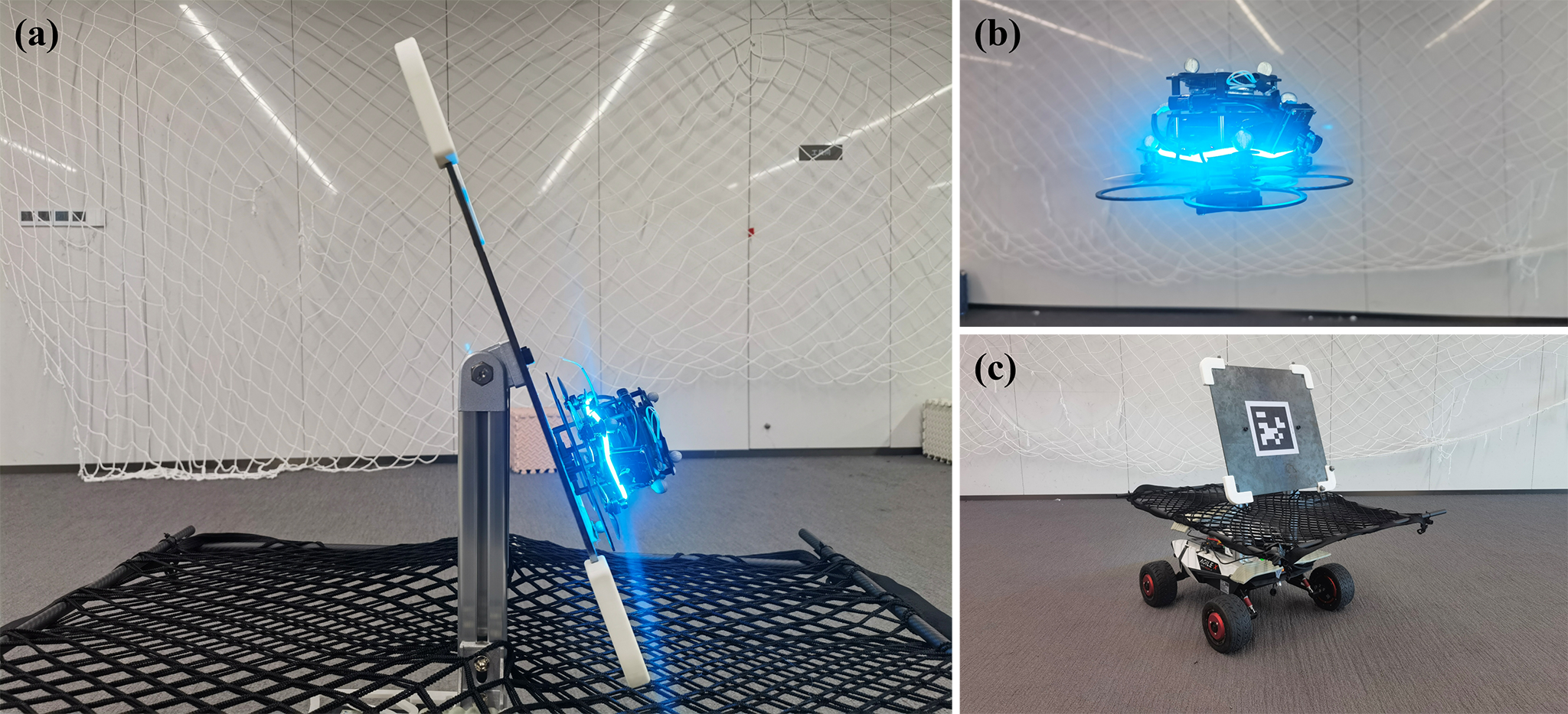}
    \caption{Hardware platforms for the vision-based perching experiments. (a) An illustration of the successful perching. (b)(c) The quadrotor and ground vehicle employed for the experiments.}
    \label{fig:Equipment_details}
    \vspace{-0.6cm}
\end{figure}

\subsection{Real-world Experiments}


We conduct a series of real-world experiments across distinct quadrotor platforms to evaluate the adaptability of PerchRL in two settings: state-based perching with a lightweight Crazyflie on a static inclined surface and onboard vision-based perching with a custom-built quadrotor on moving inclined platforms.

\subsubsection{State-Based Perching with a Crazyflie}

We first deploy the state-based policy on a Bitcraze Crazyflie 2.1 to preliminarily validate the proposed framework under aggressive flight conditions. The tested Crazyflie weighs 31.1 g and has a thrust-to-weight ratio of approximately 1.53. Accurate state estimates of both the quadrotor and the platform are provided by an NOKOV motion capture system at 100 Hz and transmitted to an offboard computer, where a state-based policy is deployed for static inclined perching experiments. The output CTBR commands are sent to the Crazyflie and executed by the onboard low-level PID controller for real-time control of the quadrotor. In this experiment, Velcro is used to secure the quadrotor to the platform after perching. As shown in Fig.~\ref{fig:crazyflie}, the Crazyflie successfully perches on a static $90^\circ$ inclined surface by rapidly approaching the target and aligning its body $z$-axis with the surface normal before contact. This successful deployment validates the adaptability of the proposed framework and provides a preliminary basis for the subsequent vision-based perching experiments.

\subsubsection{Vision-Based Perching with a Custom-Built Quadrotor}

Following the preliminary validation, we further conduct real-world vision-based closed-loop experiments to comprehensively evaluate PerchRL on moving inclined platforms. As shown in Fig.~\ref{fig:Equipment_details}, the custom-built quadrotor weighs 610.2 g and has a thrust-to-weight ratio of 2.26. It is equipped with a PX4 flight controller and an NVIDIA Jetson Orin NX for onboard policy inference at 100 Hz, with an inference latency below 1.8 ms. A monocular global-shutter camera ($1280 \times 720$ resolution, $82.96^\circ \times 52.90^\circ$ FOV) is mounted with an approximate $15^\circ$ downward tilt for visual sensing. The visual perception module is built upon the CUDA-accelerated Isaac ROS AprilTag detector. A Scout Mini carrying a \(42\,\mathrm{cm}\times42\,\mathrm{cm}\) ferromagnetic board with adjustable inclination serves as the moving platform, while a downward-facing magnet enables attachment upon perching. In addition, a NOKOV motion capture system provides accurate odometry for both the quadrotor and the ground vehicle. The quadrotor’s odometry is provided to the policy as its state input, whereas the ground vehicle’s measurements are used only for evaluation and are unavailable to the policy.

We design six scenarios spanning three difficulty levels, namely simple, normal, and hard, with two scenarios at each level. Specifically, the three difficulty levels cover stationary perching near the perception boundary, dynamic perching under moderate platform motion, and challenging perching under rapid, irregular, or time-varying motion. Due to the page limit, in this section, we focus on the first normal scenario (denoted as \textit{\textbf{Normal-I}}) and the first hard scenario (denoted as \textit{\textbf{Hard-I}}), while the complete set of six scenarios and further experimental details are provided in the supplementary video.

\begin{figure}[t]
    \centering
    \includegraphics[width=\linewidth]{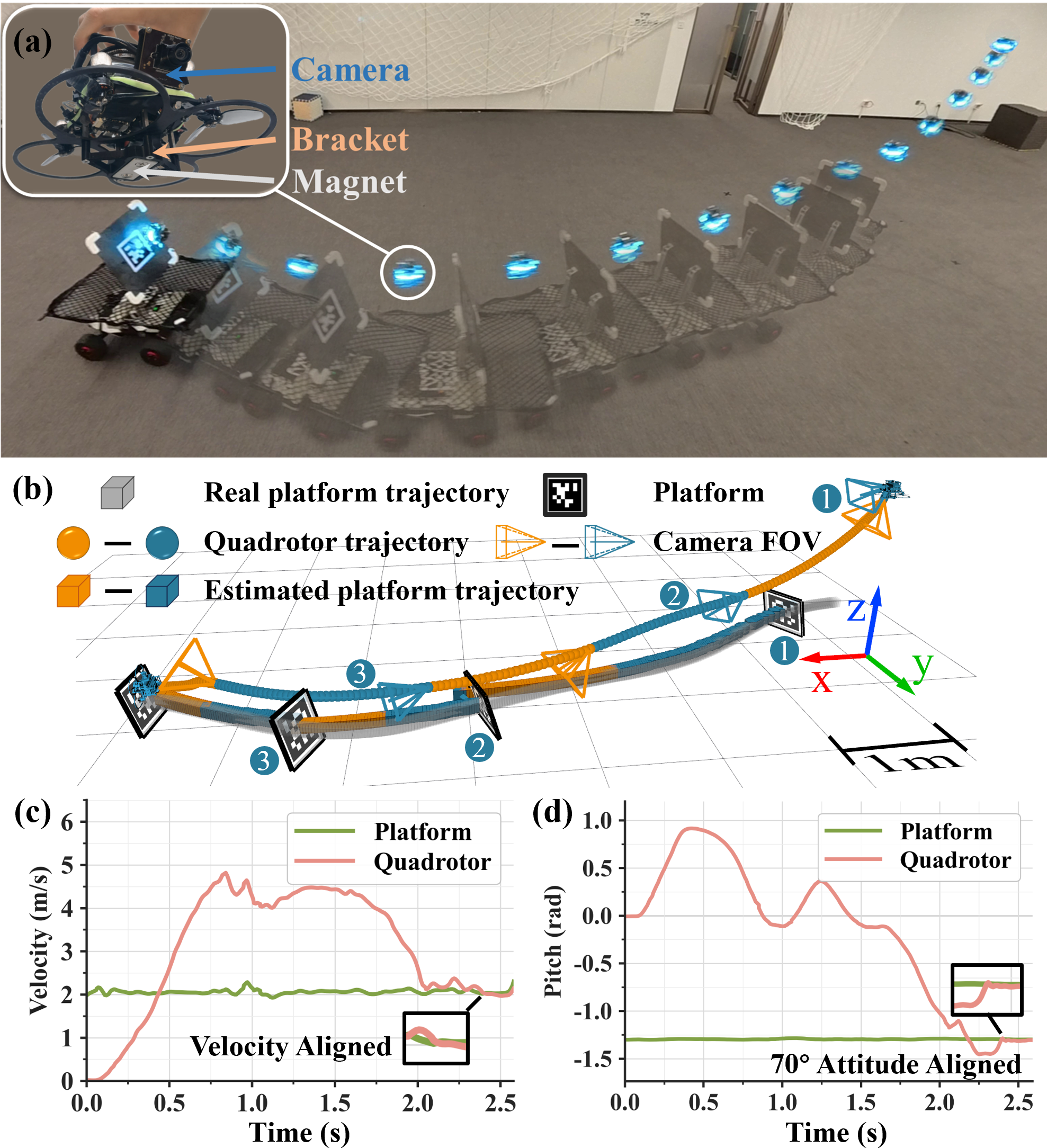}
   \caption{Illustration of the \textit{\textbf{Normal-I}} experiment. (a) Snapshot of the whole process. The quadrotor perches on the platform, which follows a spline trajectory at approximately $2.0\,\mathrm{m/s}$ with a $70^\circ$ inclination, and the inset shows the key hardware components of our custom-built quadrotor. (b) 3D trajectories of the quadrotor and platform. (c)(d) Velocity and pitch angle curves of the quadrotor and platform.}
   \label{fig:normal1_exp}
\end{figure}

\textit{\textbf{Normal-I Experiment:}} As shown in Fig.~\ref{fig:normal1_exp}(a), the experiment starts with the drone hovering, while the platform enters its FOV from the rear and follows a randomly generated spline trajectory. After accumulating sufficient valid visual detections, the drone accelerates to catch up and complete the perching task. Fig.~\ref{fig:normal1_exp}(b) shows the full trajectories, where visual perception is repeatedly lost and recovered. Each numbered marker indicates a moment when visual perception is re-established. With encoded visual perceptual reliability, the trained policy enables the drone to proceed with the perching maneuver while preventing prolonged visual loss. Ultimately, the drone successfully perches on the inclined platform with precise state alignment, as shown in Fig.~\ref{fig:normal1_exp}(c)(d).

\textit{\textbf{Hard-I Experiment:}} Fig.~\ref{fig:hard1_experiment} illustrates a more challenging scenario. Specifically, the platform’s desired velocity varies sinusoidally within the range of $[1.0, \,2.5]\,\mathrm{m/s}$. Due to the imperfect performance of the onboard ground vehicle controller, the actual velocity slightly exceeds the predefined bounds and exhibits additional fluctuations. Notably, the above velocity modulation differs significantly from the random motion used in simulation. However, our policy is able to achieve strong zero-shot generalization to this task, leading to successful perching. 

Table~\ref{tab:real_experiment_results} summarizes the relevant performance metrics across different scenarios. Notably, we specifically conduct repeated trials in \textit{\textbf{Hard-I}} and \textit{\textbf{Hard-II}} to obtain more reliable statistics under such challenging conditions. Their reported performance metrics are averaged over multiple trials, while success rates and failure modes are additionally summarized. Overall, the experimental results demonstrate that PerchRL enables vision-based perching on inclined platforms under rapid and irregular motion, with strong generalization across diverse and unseen platform motions.



 \begin{table}[t]
\centering
\caption{Real-world experimental results under different scenarios}
\label{tab:real_experiment_results}
\setlength{\tabcolsep}{3.0pt}
\renewcommand{\arraystretch}{1.08}
\resizebox{\linewidth}{!}{%
\begin{tabular}{@{}ccccccc@{}}
\toprule
\textbf{\makecell[c]{Scenario}} &
\textbf{\makecell[c]{Platform\\speed (m/s)}} &
\textbf{\makecell[c]{Inclination\\angle ($^\circ$)}} &
\textbf{\makecell[c]{Planar\\error (m)}} &
\textbf{\makecell[c]{Alignment\\error ($^\circ$)}} &
\textbf{\makecell[c]{Impact velocity\\$(v_N,v_T)$ (m/s)}} &
\textbf{\makecell[c]{Success/Trials\\(P/C/M failures)}} \\
\midrule
Simple-I
& 0.00
& 65.00
& 0.06
& 0.87
& $(-0.60,\,0.04)$
& -- \\
Simple-II
& 0.50--2.35
& 45.00
& 0.23
& 2.07
& $(-0.63,\,0.17)$
& -- \\
\midrule
Normal-I
& 2.00
& 70.00
& 0.21
& 12.12
& $(-0.39,\,0.31)$
& -- \\
Normal-II
& 2.00
& 70.00
& 0.09
& 11.94
& $(-0.37,\,0.08)$
& -- \\
\midrule
Hard-I
& 0.70--2.70
& 70.00
& $0.15\pm0.05$
& $7.65\pm3.94$
& $(-0.53\pm0.17,\,0.17\pm0.06)$
& $7/10\;(1/1/1)$ \\
Hard-II
& 1.50--2.20
& 70.00
& $0.18\pm0.07$
& $8.13\pm4.08$
& $(-0.61\pm0.24,\,0.24\pm0.09)$
& $7/10\;(2/0/1)$ \\
\bottomrule
\end{tabular}%
}

\vspace{2pt}
\parbox{\linewidth}{%
\scriptsize
Note: $v_N$ and $v_T$ denote the normal and tangential
impact-velocity components. Trial outcomes report successful/total trials,
with perception-loss/contact/magnetic-attachment failures (P/C/M) in
parentheses.
}
\vspace{-12pt} 
\end{table}

\section{Conclusion}

In this work, we presented PerchRL, a learning-based framework for vision-based agile perching on inclined platforms under rapid and irregular motion. The system adaptively balances agile perching maneuvers with visibility maintenance, enabling robust perching despite intermittent visual loss. Simulation and real-world experiments demonstrate the effectiveness of jointly considering perception and control within a unified RL framework, especially when continuous visual feedback is difficult to maintain reliably. In future work, we plan to reduce the residual perception-control coupling caused by underactuated dynamics and rigid camera mounting by incorporating an active gimbal and jointly optimizing flight maneuvers and camera pointing.

\bibliographystyle{IEEEtran}
\bibliography{references}

\end{document}